\documentclass[a4paper,twocolumn]{article}
\usepackage{booktabs}
\usepackage{times}
\usepackage{epsfig}
\usepackage{graphicx}
\usepackage{amsmath}
\usepackage{amssymb}
\usepackage{subfigure} 
\usepackage{multirow}
\usepackage[pagebackref=true,breaklinks=true,colorlinks,bookmarks=false]{hyperref}
\date{}
\usepackage{authblk}

\begin{document}

%%%%%%%%% TITLE
\title{Multi-channel Deep Supervision for Crowd Counting}

\author{Bo Wei, Mulin Chen, Qi Wang\thanks{Corresponding author}, Xuelong Li \\School of Computer Science and Center for OPTical IMagery Analysis and Learning (OPTIMAL), \\Northwestern Polytechnical University, Xi'an 710072, Shaanxi, P. R. China}
\maketitle

%%%%%%%%% ABSTRACT
\begin{abstract}
  Crowd counting is a task worth exploring in modern society because of its wide applications such as public safety and video monitoring.  Many CNN-based approaches have been proposed to improve the accuracy of estimation, but there are some inherent issues affect the performance, such as overfitting and details lost caused by pooling layers. To tackle these problems, in this paper, we propose an effective network called MDSNet, which introduces a novel supervision framework called Multi-channel Deep Supervision (MDS). The MDS conducts channel-wise supervision on the decoder of the estimation model to help generate the density maps. To obtain the accurate supervision information of different channels, the MDSNet employs an auxiliary network called SupervisionNet (SN) to generate abundant supervision maps based on existing groundtruth. Besides the traditional density map supervision, we also use the SN to convert the dot annotations into continuous supervision information and conduct dot supervision in the MDSNet. Extensive experiments on several mainstream benchmarks show that the proposed MDSNet achieves competitive results and the MDS significantly improves the performance without changing the network structure.
\end{abstract}

%%%%%%%%% BODY TEXT
\section{Introduction}
Crowd counting is a visual task that aims to count the number of people contained in given static images. The continuous growth of the population and urbanization have led to more and more gatherings such as public demonstrations, political rallies and sports events. As a result, crowd counting has attracted remarkable attention in recent years because of its wide applications such as public safety, video surveillance and traffic control.

According to \cite{2017A}, crowd counting approaches can be divided into three categories: Detection-based methods, Regression-based methods and CNN-based methods. Detection-based methods\cite{2005Histograms}\cite{2005Detection}\cite{2006Detecting}\cite{extra1}\cite{Dectect3}\cite{D4} use sliding windows to detect and count individuals in static images and regression-based methods\cite{ChanLV08}\cite{ChenLGX12}\cite{Regression1} aim at creating a mapping between features and the total number of people. Though these approaches have achieved good results in some scenarios, they have a poor generalization. In recent years, with the development of deep learning, CNN-based methods have achieved higher accuracy and better generalization performance than other methods. Many CNN-based methods\cite{WangZYLC15}\cite{MCNN}\cite{CrowdNet}\cite{extra3}\cite{IJCV} have been proposed to meet the challenge of crowd counting and improve the results, such as various scales and uneven population distribution. However, there are still some inherent issues in CNN-based methods severely affect the performance of networks which are rarely discussed. 

The first issue is the lack of crowd data. It has been proved that the number of network parameters is positively correlated with the final performance and generalization ability of the network. But overfitting easily occurs if the size of the dataset is not big enough to match the size of the network. The usual approaches to solve this issue are simplifying network structure and augmenting the datasets. But reducing the parameters of CNN will weaken the ability of feature extraction and expression. It's also difficult to expand new crowd data because manual annotation takes a lot of manpower. The second issue is lost of details of feature maps caused by pooling operations. Crowd counting is a task sensitive to details, especially in extremely dense scenes. Though the pooling layers extract high-level features which have inherent invariance, they reduce the resolution of the image and make some pixels in feature maps lost. Some methods employ upsampling operations, such as interpolations and deconvolutions, to improve the resolution, but they can hardly compensate for the impact of details lost. The third one is the effect of gaussian kernel used in the generation of groundtruth. The initial groundtruth of the crowd image is a binary image contains dot annotations. For the convenience of training, many methods\cite{WangZYLC15}\cite{LempitskyZ10} use gaussian convolution to make the dot cover the person and convert the binary image into a continuous density map. But because of the occlusions and the various scales, the kernels are hard to decide in different scenarios. \cite{MCNN} proposed an adaptive method to get the kernel size based on the distance between individuals. However, it causes large errors in sparse scenes.

To tackle the problems mentioned above, in this paper, we propose a novel network call Multi-channel Deep Supervision Network (MDSNet), which introduces an effective supervision framework called Multi-channel Deep Supervision (MDS). The MDSNet consists of an auxiliary network called SupervisionNet (SN) and a Density Map Estimation (DME). The SN converts the traditional single-channel groundtruth into multi-channel supervision information and employs the channel-wise attention mechanism to select the more important channels. It's an effective way to avoid overfitting by making the best use of existing groundtruth. With the help of proposed supervision information, we conduct MDS on the DME to supplement the details lost by pooling. Different from traditional deep supervision methods which use true labels\cite{DSN}\cite{True3}\cite{True4} or related labels\cite{DSPAMI}\cite{Related2} as the intermediate groundtruth, our MDS focuses on the high-level features in different channels and conducts supervision on channels of hidden layers. Besides the traditional density map supervision, we use SN instead of the hand-crafted gaussian kernel to convert the dot annotations into continuous supervision information and conduct the dot supervision on our network. To sum up, the main contributions of this paper are:
\begin{itemize}
	\item We analyze the limitations of traditional deep supervision and propose an improved framework called MDS. The MDS imposes additional restrictions on the channels of feature maps directly, which alleviate the overfitting and supplement the details lost in forward propagation.
	
	\item We design an auxiliary network called SN to convert the existing groundtruth into abundant supervision information. Besides traditional density map supervision, we conduct dot supervision on our network with the help of SN.
	
	\item Extensive experiments show that our MDSNet has achieved a competitive performance on three challenging datasets in terms of traditional density map supervision and dot supervision. Moreover, the ablation experiments show that the MDS significantly improves the performance without changing the network structure.
\end{itemize}
%-------------------------------------------------------------------------
\section{Related Work}
\begin{figure*}[t]
	\centering
	\includegraphics[scale=0.45]{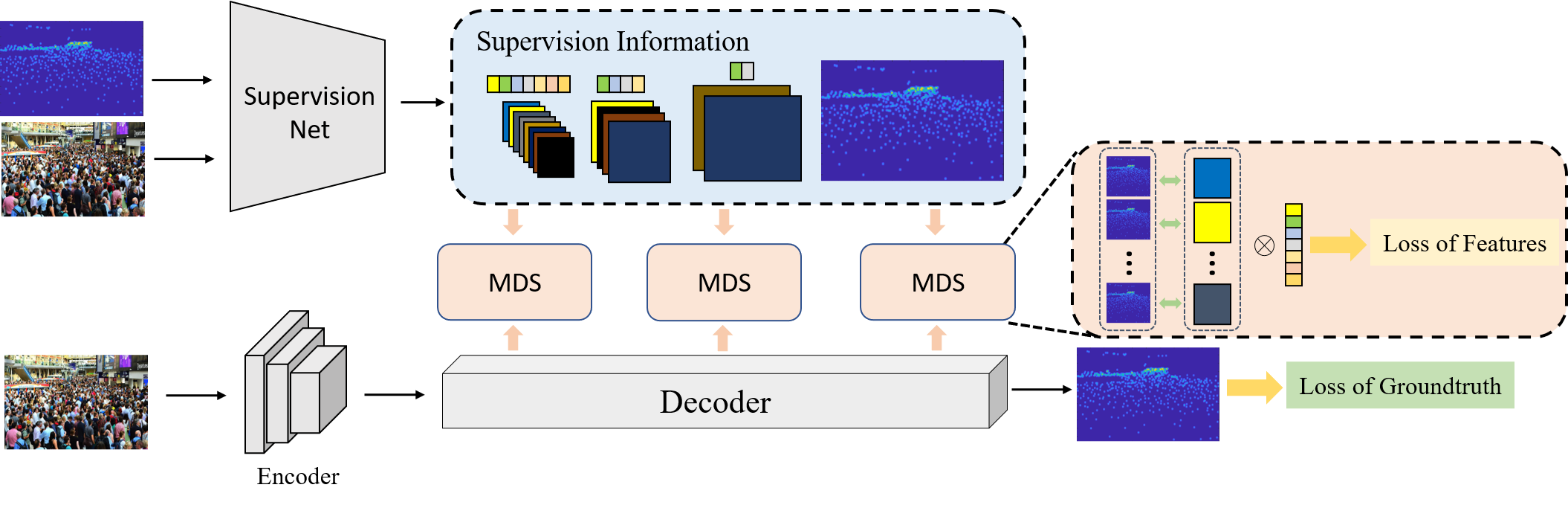} 
	\caption{Overview of the proposed MDSNet architecture. It consists of Density Map Estimation (DME) and SupervisionNet (SN). The SN is an auxiliary network which is pretrained to extract the supervision information of the groundtruth. DME is an encoder-decoder architecture. MDS is conducted on the decoder of DME.}
	\label{fig1}
\end{figure*}
Most of crowd counting methods are CNN-based methods\cite{coushu1}\cite{coushu2}\cite{coushu3}\cite{coushu4}\cite{coushu5}\cite{coushu6} in recent years. In this section, some of them are briefly reviewed. In addition, several deep supervision methods\cite{coushu7}\cite{coushu8} which are related to our proposed MDSNet are also described.

\textbf{Methods of extracting various features.} With the development of computing power, deep learning has been proved effective in many computer vision tasks, including crowd counting. These methods\cite{MCNN}\cite{NWPUdataset}\cite{Useless1}\cite{Useless2} use unique network structures to solve different challenges, such as uneven scales and population distributions. For various scales in images, Zhang \emph{et al.} proposed a multi-column network (MCNN)\cite{MCNN} which has three branches with different receptive fields. The purpose of MCNN is combining features to adapt to different scales. Sam \emph{et al.} proposed a switching CNN architecture\cite{Switching-CNN} based on MCNN, which uses a switching layer to choose the optimal estimation CNN according to the scale of the input image. In SANet\cite{SANet}, scale aggregation modules are employed to extract and combine features in different hiddle layers. Li \emph{et al.} proposed a single-column network called CSRNet\cite{CSRNet}, which uses dilated convolutions to enlarge the receptive field without increasing the parameters of the  network. 

For uneven population distributions, Sindagi \emph{et al.} proposed a parallel network architecture\cite{CMTL} to improve the accuracy of estimation with a high-level prior. DecideNet\cite{DecideNet} employed a regression block and a detection block to estimate dense crowd and sparse crowd respectively. Sindagi and Vishal proposed the CP-CNN\cite{CP-CNN} to introduce the contextual information by two extra branches called Global Context Estimator and Local Context Estimator.   
Obviously, though these methods have achieved good performances, most of them added extra branches to extract various features, which made the network architecture redundant.

\textbf{Methods of enlarging dataset} The lack of labeled data is an enormous challenge in crowd counting because the crowd images are hard to get and the annotation is a time-consuming and labor-intensive work. To overcome this issue, Zhang \emph{et al.} proposed a data-driven method\cite{Cross-scene} to adopt the network to a new scene by using existing data that similar to the target scene. Kang \emph{et al.} proposed a collaborative learning method\cite{KangDC16} to estimate density with the help of images and side information, such as camera angle and height. But this method has a poor generalization ability and the side information is expensive to obtain. In \cite{Self-supervision}, a self-supervised method was proposed to reduce the overfitting. This method leveraged ranking data cropped from unlabeled images during training. Wang \emph{et al.} proposed a domain adaptation method\cite{GTA} to generate synthetic crowd data from GTAV and created a huge synthetic crowd dataset called GCC.  

\textbf{Methods of deep supervision} The concept of deep supervision was first proposed in Deeply Supervised Nets(DSN)\cite{DSN}. In DSN, the groundtruth label was used to supervise the middle layers of the network. In \cite{DSN2}, deep supervision was applied to edge detection to learning important hierarchical representations. In GoogleNet\cite{GoogleNet}, some extra branches called auxiliary classifiers generated intermediate results. Li \emph{et al.} proposed an improved deep supervision method\cite{DSPAMI} which used intermediate concepts to supervised hidden layers of network and proved the effectiveness of deep supervision in generalization. However, the intermediate concepts related to the final task need to be designed manually. 

\textbf{Methods of groundtruth generation} Convolving the dot map with a Gaussian kernel into the continuous density map is the mainstream method in crowd counting. To tackle the scale variation, Zhang \emph{et al.} employed the Gaussian kernel with an adaptive bandwidth to generate groundtruth\cite{MCNN}. The bandwidth is decided by the distances between the target and its neighbors. Though this method improves the performance in the dense crowds, it leads a large error in sparse scenes. Wan \emph{et al.} proposed an adaptive generation method\cite{Wan} which uses an auxiliary network to generate groundtruth instead of hand-crafted kernels. \cite{Bayesian} uses a novel loss function called Bayesian loss to estimate the count expectations around the dot annotations, but this method still needs to design the hyper-parameters experientially.

\section{Method}
In this section, we first introduce the MDS and the generation of supervision information. Then we revisit the network architecture of MDSNet, as shown in Fig\ref{fig1}. Furthermore, we explore the feasibility of dot supervision.
\subsection{Multi-channel Deep Supervision}
\label{3.1}
We propose a novel supervision method called MDS. The MDS uses the supervision information corresponding to different channels to conduct the channel-wise deep supervision on the feature maps, which imposes constraints on the hidden layers directly. It's an effective way to tackle the overfitting and the lost of details in forward propagation.

\begin{figure}[t]
	\centering
	\subfigure[]{
		\includegraphics[scale=0.5]{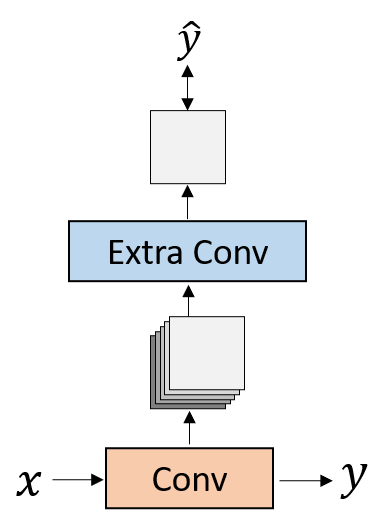}
		\label{fig2a}
	}
	\subfigure[]{
		\includegraphics[scale=0.5]{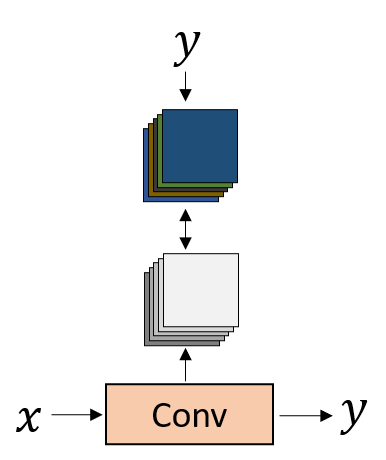}
		\label{fig2b}
	}     
	\caption{(a) and (b) are the illustrations of traditional deep supervision method and MDS, respectively.}
\end{figure}

We first use a simple example to describe the principle of the MDS. Similar to \cite{DSPAMI}, considering a small three-layer network: $y = \sigma \left( {{w_3}\sigma \left( {{w_2}\sigma \left( {{w_1}x + {b_1}} \right) + {b_2}} \right) + {b_3}} \right)$ where $w$ and $b$ are weights need to train, and $\sigma$ is ReLu activation $\sigma\left(x\right)=\max \left( {x,0} \right)$. The overfitting is easily aggravated if the size of dataset is too small to impose enough constraints on the parameters of network. Now we have a dataset $\left\{ {\left( {x,y} \right)} \right\}$ which consists of the training data $\left\{ {\left( {1,0} \right),\left( {2,0} \right),\left( {3,0} \right)} \right\}$ and the test data $\left\{ {\left( {4,0} \right),\left( {5,0} \right)} \right\}$. It's obvious that model $\left( {{w_1},{w_2},{w_3},{b_1},{b_2},{b_3}} \right) = \left( {2,1,1, - 3, - 4,0} \right)$ achieves zero loss over the training data but hard to work on the test data. To tackle the overfitting and improve the generalization, the traditional deep supervision methods add extra convolutional layers to convert the multi-channel feature maps into single-channel intermediate representations. Then the true labels\cite{DSN} or the concepts with related semantic\cite{DSPAMI} are employed to supervise the intermediate representations, as shown in Fig.\ref{fig2a}. By introduce the intermediate label $\hat{y}$, the dataset is expanded to $\left\{ {\left( {1,\hat{y},0} \right),\left( {2,\hat{y},0} \right),\left( {3,\hat{y},0} \right)} \right\}$ and the forward propagation is improved as: 
\begin{equation}
\label{e1}
\left\{ \begin{array}{l} 
y = \sigma \left( {{w_3}\sigma \left( {{w_2}\sigma \left( {{w_1}x + {b_1}} \right) + {b_2}} \right) + {b_3}} \right)\\
\hat{y} = \hat{w}\sigma \left( {{w_2}\sigma \left( {{w_1}x + {b_1}} \right) + {b_2}} \right) + \hat{b}
\end{array} \right.,
\end{equation}
Where the $\hat{w}$ and $\hat{b}$ are weights of the extra convolutional layer. However, because of the $\hat{w}$ and the $\hat{b}$, the constraints are hard to work on the network directly. The parameters in the extra layer change to adapt to the constraint, but the parameters in the backbone are not effectively affected. In addition, the meaning of single-channel intermediate representation generated by the extra convolutional layer is hard to decide. It's obvious that the result is generated gradually in the network and the expressions of hidden layers should be different, so using groundtruth to conduct deep supervision is not accurate.

To tackle the limitations of traditional deep supervision, we remove the extra convolutional layers and conduct the MDS on the feature maps in the hidden layers directly, as shown in Fig.\ref{fig2b}. In other words, it removes the weights $\hat{w}$ and $\hat{b}$ in formulation\ref{e1} and imposes constraints directly on the parameters of the backbone. In addition, we convert the existing groundtruth into supervision information to conduct MDS instead of using the groundtruth as the intermediate labels directly. Compared with the true labels, the generated supervision information contains more high-level features and is better for the supervision of hidden layers in the network. 

%-------------------------------------------------------------------------
\subsection{The Generation of Supervision Information}
\label{3.2}
The performance of MDS is heavily dependent on the accuracy of supervision information. In our MDSNet, We employ an auxiliary network called SN to generate the supervision information. In fact, most architectures of crowd counting methods are encoder-decoder. The encoder extracts the features and the decoder generates the density maps. The purpose of SN is to propose the information which is useful to the generation of density maps.

The SN is an encoder-decoder with two inputs. The encoder has two branches with the same architectures, which both consist of four convolutional layers. To simplify the structure, we use the convolutional layer with the stride of 2 instead of the pooling layer. The decoder employs deconvolutional layers to generate the groundtruth of the original size. The encoder-decoder can be regarded as a process of decomposition and reconstruction of groundtruth. Besides, the crowd image is also input to propose context information to the network. After pretrained the SN, we extract the feature maps of the decoder as the supervision information of the backbone.

\begin{figure*}[t]
	\centering
	\includegraphics[scale=0.35]{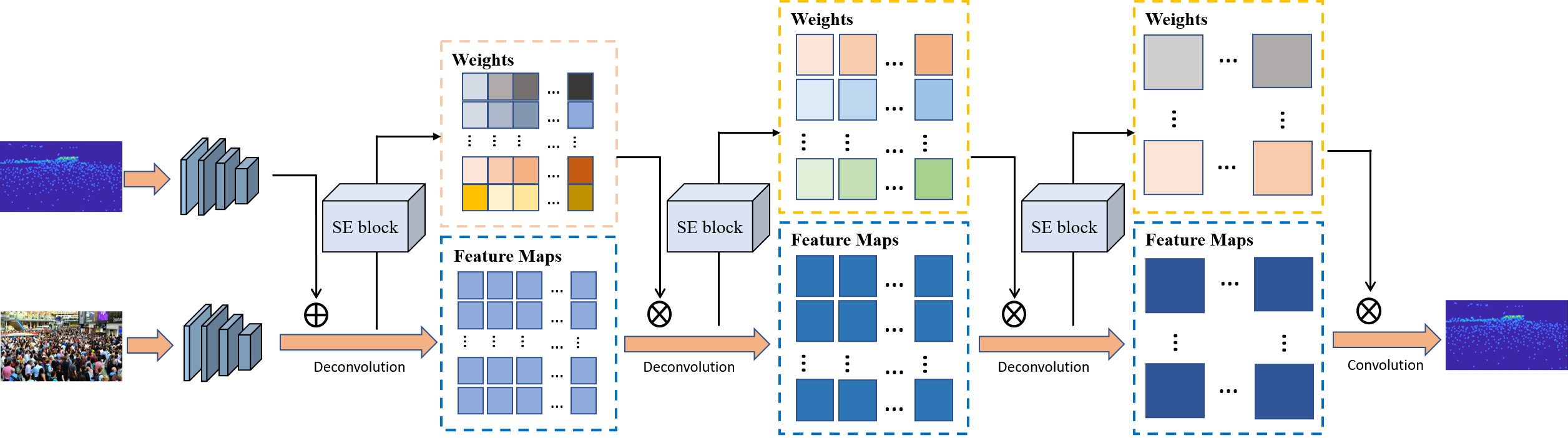}
	\caption{The detailed architecture of SN.}
    \label{fig3}
\end{figure*}

Because of the overfitting, there is a lot of redundant information exists in the feature maps of decoder. It's quite possible that using the useless information to supervise the backbone will degrade the performance. To select the redundant information, we introduce the attention mechanism to the SN. We assume that the channels which have greater contributions to generate the groundtruth have more useful information. To select the target channels, the channel-wise attention mechanism is employed to propose different weights of different channels. Similar to \cite{SEblock}, we employ the Squeeze-and-Excitation block (SE block) to conduct unsupervised channel-wise attention. In the decoder of SupervisionNet, the feature maps are input to the SE block to generate various weights. Then channel-wise multiplication is performed between feature maps and weights. By training the SupervisionNet, the weights generated by SE blocks correspond to the importance of the different channels. The architecture of SupervisionNet is shown in Fig.\ref{fig3}. 
%\vspace{-1.5em}

\subsection{Network Architecture}
Fig.\ref{fig1} shows the architecture of our MDSNet, which consists two subnetworks, \emph{i.e.} SN and DME. The architecture of SN has been described in Sec.\ref{3.2}. It's worth noting that the proposed supervision framework is generic and easy to apply to many encoder-decoder DMEs. In the experiments, for the sake of concreteness and implementation, we employ the first ten convolutional layers of pre-trained VGG-16\cite{VGG} as the encoder of DME because of its strong learning ability. To conduct the MDS in the DME, we design three sets of layers in the decoder and every set consists of a convolutional layer and a deconvolutional layer, which restore the image to the original resolution. The MDS is implemented in every set of layers to supplement the details of feature maps and alleviate the overfitting. After the three sets of layers, we add two convolutional layers to generate the final density map. The training is divided into two stages. The SN is trained first to learn how to generate the density maps and then we extract the information in decoder of SN to train the DME.
%\vspace{-0.5em}

\subsection{Dot Supervision}
\label{3.4}
Besides the traditional density map supervision, we also conduct dot supervision in our MDSNet. Dot map is a binary image with the value 0 and 1. The 1 means a head in this pixel and the 0 means background or other parts of body. Using the binary image training the network is easy to lead the vanishing gradient problem because the image is sparse. Another disadvantage is the dot map doesn't contain the context information because the other parts of body in the dot maps are recognized as the background, which makes the estimation more difficult. To tackle the two issues, many methods employ gaussian convolution to convert the dot annotations into continuous density maps. In our MDSNet, we obtain continuous supervision information from SN instead of hand-crafted gaussian kernel, which is employed to supervise the hidden layers to alleviate the vanishing gradient problem. Because of the input dot image and crowd map, the generated supervision information contains localization information and context information, which help generate accurate results.

\section{Experiments}
\subsection{Evalution metrics}  
\label{4.1}
The Mean Absolute Error(MAE) and the Mean Squared Error(MSE) are usually adopted in crowd counting methods. The formulations are defined as follows:
\begin{equation}
	\label{e2}
	{MAE{\rm{ = }}\frac{{\rm{1}}}{N}\sum\limits_{i = 1}^N {\left| {{Y_i} - {{\hat Y}_i}} \right|} ,}
\end{equation}
\begin{equation}
	\label{e3}
{MSE{\rm{ = }}\sqrt {\frac{{\rm{1}}}{N}\sum\limits_{i = 1}^N {{{\left\| {{Y_i} - {{\hat Y}_i}} \right\|}^2}} } ,}
\end{equation}
where $N$ is the number of test samples. $Y_i$ is the groundtruth of crowd counting and ${\hat Y_i}$ represents the predicted number of $i$th image, respectively. 

\subsection{Implementation details}
\label{4.2}
\textbf{Data} We adopt a data augmentation in the training process. When generating the training set, we crop the patches of 400×400 from the original images at random locations. In addition, random horizontal flipping and gamma transformation are also employed to increase the diversity of data.

\textbf{Groundtruth generation} The density map ground truth is generated according to, which is defined as:
\begin{equation}
	\label{e4}
{G^{GT}}\left( x \right) = \sum\limits_{i = 1}^N {\delta \left( {x - {x_i}} \right) \times {G_\sigma }\left( x \right),}
\end{equation}
Where the ${{G_\sigma }}$ is the Gaussian kernel with the standard deviation of $\sigma$. $\delta \left( {x - {x_i}} \right)$ is the dot annotation. For the datasets which are relatively dense, such as ShanghaiTech Part A\cite{MCNN}, UCF\_CC\_50\cite{UCF_CC_50} and UCF-QNRF\cite{UCF-QNRF}, the $\sigma$ is set to 5. For ShanghaiTech Part B dataset, we set the $\sigma$ as 15.

\textbf{Objective function} We use MSELoss as the loss of groundtruth, which is a widely-used objective function in crowd counting. The formulation is defined as:
\begin{equation}
	\label{e5}
{{L_G} = \frac{1}{N}\sum\limits_{i = 1}^N {{{\left\| {O\left( {{X_i},\Theta } \right) - {G_i}} \right\|}^2}},}
\end{equation}
where N is the number of samples in batch. ${O\left( {{X_i},\Theta } \right)}$	is the output of network with parameters $\Theta$ and input $X_i$. $G_i$ is the groundtruth of $i$th image. 

As mentioned in section \ref{3.1}, we implement MDS in the training process. As shown in Fig.\ref{fig1}, loss of features is generated by the feature maps and the supervision information. The loss of features is defined as:
\begin{equation}
	\label{e6}
{L_{F}} = \frac{1}{N}\sum\limits_{i = 1}^N {\sum\limits_{n = 1}^M {{\beta _n}\sum\limits_{c = 1}^{{C_n}} {W_c^n{{\left\| {O\left( {{X_i},\Theta _c^n} \right) - G_c^n} \right\|}^2}} },} 
\end{equation}
where ${\Theta _c^n}$ are the parameters of $c$ the channel in $n$ node. $G_c^n$ is the supervision information and $W_c^n$ is the corresponding weight generated by channel-wise attention. $C_n$ is the total number of channels in $n$th node and $M$ is the number of supervision nodes, which is set to three in our experiments. The $\beta _n$ is the weight of different nodes. Intuitively, the node closer to the end of network is more important, so we employ an Increase strategy to decide the $\beta _n$. The Increase strategy is defined as:
\begin{equation}
\label{e7}
{\beta _n} = \frac{1}{{{2^{M - n}}},}
\end{equation}
where $\beta_n$ increases in an equal proportion that the weight of current node is half of the next node. The value deponds on the total number of supervision nodes.

The final loss function is a combination of loss of groundtruth and loss of features, which is defined as:
\begin{equation}
\label{e8}
{L = {L_G} + \alpha {L_{F}},}
\end{equation}
where $\alpha$ is used to balance the multi-channel losses and the final loss. In our experiments, we set $\alpha$ to 0.05. 

To conduct dot supervision, inspired by \cite{Bayesian}, we regard the count regression as probability estimation. Crowd counting is a special regression task that the value of pixel is 0 or 1. The result of probability estimation in a single pixel represents the probability that there is a person. The total counting could also be obtained by summing the pixel in probability map because the value of pixel is the mathematical expectation of count. To estimate the probability, we employ the CrossEntropy Loss, which is defined as:
\begin{equation}
\label{e10}
{L_{Dot}} =  - \frac{1}{N}\sum\limits_{i = 1}^N {\left( {{y_i}{\rm{In}}{{\hat y}_i} + \left( {1 - {y_i}} \right){\rm{In}}\left( {{\rm{1 - }}{{\hat y}_i}} \right)} \right),} 
\end{equation}
where the $y$ is the goundtruth and $\hat y$ is the estimated value. The CrossEntropy Loss is only used to train the output and MSELoss is still used to conduct deep supervision.
\subsection{Datasets}
\label{4.3}
We conduct experiments on several commonly-used datasets, including ShanghaiTech\cite{MCNN}, UCF-QNRF\cite{UCF-QNRF} and UCF\_CC\_50\cite{UCF_CC_50}.

\textbf{ShanghaiTech dataset} The ShanghaiTech dataset is created by Zhang \emph{et al.}, which contains 1198 images which 330,165 annotated heads. This dataset is divided into two parts: 482 images in Part A, which are crawled from the Internet randomly, and 716 images in Part B, which are taken from the street in Shanghai. 

%Table\ref{table1} shows the performances of the proposed MDSNet and other several methods, including MCNN, Switching-CNN, CSRNet, CAN and so on. Compared with other methods, the proposed MDSNet achieves state-of-the-art performance on ShanghaiTech Part A and competitive performance on ShanghaiTech Part B. Specifically speaking, the MDSNet achieves the best performance in terms of MAE(1.7-point improvement) and MSE(0.2-point improvement) on Part A. On the Part B, the MDSNet achieves the third-best MAE of 8.4 and MSE of 14.3. The results on Part A have proved the effectiveness of MDSNet in dense areas. But in the sparse Part B dataset, the grountruth is hard to convert into useful supervision information, which has affected on the performance of MDSNet.

\textbf{UCF\_CC\_50 dataset} The UCF\_CC\_50 dataset is proposed by Idrees \emph{et al.}, which contains only 50 images but shows a lot of challenges. The distribution of people count in this dataset is vastly different, which ranges from 94 to 4543. The total number of labeled individuals is 63,075 and the average labeled number per image is 1280. The commonly-used way of UCF\_CC\_50 is five-fold cross-validation protocol, which is also performed in our experiments. 

%The estimation errors of the proposed MDSNet and some other state-of-the-art methods are listed in table\ref{table2}. Our MDSNet achieves the best MAE of 200.1 and the second-best MSE of 310.6. The results have prove that our MDSNet has good robustness in extremely crowded scenes. 
\textbf{UCF-QNRF dataset} The UCF-QNRF dataset is a large-scale crowd counting dataset which contains 1,535 images with 1,251,642 annotations. The people count ranges from 49 to 12,865. In the dataset, 1,201 images are assigned to the training set and the rest to the test set. The density of crowd and resolutions of images have extreme variations, which make this dataset challenging for deep learning methods.

\subsection{Performance Comparisons}
%Table\ref{table3} shows the results of our MDSNet and other state-of-the-art methods. Our method achieves the best performance in MAE(6-point improvement) and MSE(12-point improvement).
Table \ref{table1} shows the results of our proposed method and the state-of-the-art methods on the datasets introduced in Sec.\ref{4.3}. In the table, the Baseline is the same structure without multi-channel supervision. MDSNet-D means the MDSNet with dot supervision. Compared with other methods, our MDSNet achieves a competitive performance in these benchmarks, especially in the UCF\_CC\_50. The MDSNet-D also achieves promising performance compared with BL, which conducts dot supervision as well. Compared with Baseline, with the help of MDS, the performance is improved significantly. It's worth noting that our MDSNet is a simple single-column network because the SN is only used in the training process. The competitive performance has proved the effectiveness of MDS.

\begin{figure*}[t]
	\centering
	\includegraphics[scale=0.7]{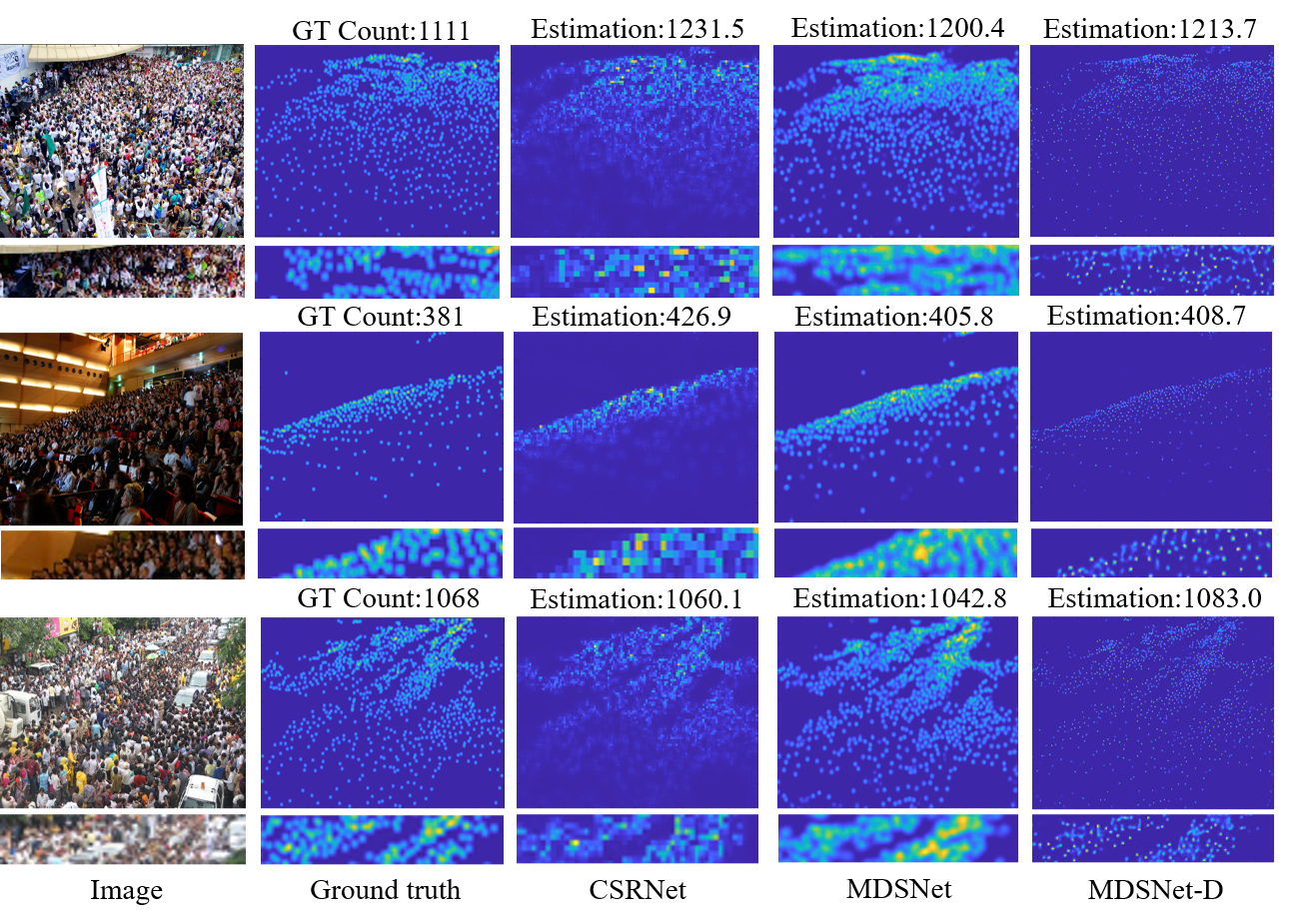}
	\caption{The visualized results. Compared with the CSRNet, the results of MDSNet have higher resolution and accuracy. The rectangle images are the details of congested areas in the shown images. It's obvious that the results of MDSNet-D show more localization information.}
	\label{fig4}
\end{figure*}

The visualization results of the networks are shown in Fig.\ref{fig4}. Compared with the results fo CSRNet, our results have high resolutions and quality. In addition, besides the promising counting accuracy, the results of MDSNet-D show the localization of people in crowd, which are more conducive to practical applications than the traditional density maps.

\begin{table*}[h]
	\caption{Evaluations on three widely-used crowd counting datasets. Baseline has the same structure as the DME in our MDSNet. MDSNet-D is the MDSNet supervised by dot annotations directly.}
	\label{table1}
	\begin{center}
		\setlength{\tabcolsep}{4mm}{
			\begin{tabular}{c|cc|cc|cc|cc}
				\toprule[2pt]
				\multirow{2}{*}{\begin{tabular}[c]{@{}c@{}}Datasets\\ Methods\end{tabular}} & \multicolumn{2}{c|}{SHA} & \multicolumn{2}{c|}{SHB} & \multicolumn{2}{c|}{UCF-QNRF} & \multicolumn{2}{c}{UCF\_CC\_50} \\ 
				& MAE         & MSE        & MAE         & MSE        & MAE           & MSE           & MAE             & MSE            \\ \midrule[1.5pt]
				MCNN\cite{MCNN}                                                                        & 110.2       & 173.2      & 26.4        & 41.3       & 277           & 426           & 377.6           & 509.1          \\ 
				CMTL\cite{CMTL}                                                                        & 101.3       & 152.4      & 20.0        & 31.1       & 252           & 514           & 322.8           & 341.4          \\ 
				Switching-CNN\cite{Switching-CNN}                                                               & 90.4        & 135.0      & 21.6        & 33.4       & 228           & 445           & 291.0           & 404.6          \\
				PCCNet\cite{PCCNet}                                                                      & 73.5        & 124.0      & 11.0        & 19.0       & 149           & 247           & 240.0           & 315.5          \\ 
				ACSCP\cite{ACSCP}                                                                       & 75.7        & 102.7      & 17.2        & 27.4       & -             & -             & 291.0           & 404.6          \\ 
				CP-CNN\cite{CP-CNN}                                                                      & 73.6        & 106.4      & 20.1        & 30.1       & -             & -             & 295.8           & 320.9          \\ 
				CSRNet\cite{CSRNet}                                                                      & 68.2        & 115.0      & 10.6        & 16.0       & -             & -             & 266.1           & 397.5          \\ 
				DUBNet\cite{DUBNet}                                                                      & 64.6        & 106.8      & 7.7         & 12.5       & 105.6         & 180.5         & 243.8           & 329.3          \\ 
				CAN\cite{CAN}                                                                         & 62.3        & 100.0      & 7.8         & \textbf{12.2}      & 107           & 183           & 212.2           & 243.7          \\ 
			    TEDNet\cite{TEDNet}                                                                  & 64.2        &109.1        &8.2           &12.8               &113           &188        & 249.4    &354.5 \\
				SFCN\cite{GTA}                                                                       & 64.8        & 107.5      & \textbf{7.6}        & 13.0       & 102.0       &171.4           &214.2          &318.2             \\
				BL\cite{Bayesian}                                                                          & 62.8        & 101.8     & 7.7         & 12.7       & \textbf{88.7}          & \textbf{154.8}         & 229.3           & 308.2          \\ \midrule[1.5pt]
				Baseline       & 70.2     & 114.6   & 10.5      &17.4     & 115.2    &189.6       &280.1           &411.5              \\
				MDSNet           & 63.2       & 107.0      & 8.0           &13.2          &99.1             &168.0             &203.2      &\textbf{260.8}       \\
				MDSNet-D         & \textbf{60.6}        & \textbf{95.8}       & 7.8         & 12.3       &93.6              &155.1             &\textbf{200.1}        &262.8      \\ \bottomrule[2pt]
		\end{tabular}}
	\end{center}
\vspace{-1.5em}
\end{table*}

\subsection{Ablation Study}

To demonstrate the effectiveness of our MDSNet, we conduct the ablation experiments on ShanghaiTech Part A dataset\cite{MCNN}. The ablation experiments consist of two parts: the number of superivisons and the effect of channel-wise attention mechanism in SN.

\textbf{The number of supervisions} As mentioned above, conducting deep supervision on the hiddle layers of network in training process is conducive to the final performance. Table \ref{table3} shows the results of MDSNet with different numbers of supervision nodes. The Baseline-x means x nodes are implemented supervision in the Baseline. From the table, increasing the number of supervision improves the performance of the model.

\textbf{The effect of channel-wise attention} To prove the effectiveness of the channel-wise attention mechanism in SN, we conduct a set of comparative experiments. We design another strategy for assigning the weights of different channels of supervised feature maps called Equal Weights(EW). Every channel has the same weight in training process in EW. The loss function of EW is defined as:
\begin{equation}
	\label{e9}
{L_{F}} = \frac{1}{N}\sum\limits_{i = 1}^N {\sum\limits_{n = 1}^M {{\beta _n}\sum\limits_{c = 1}^{{C_n}} {\frac{{{{\left\| {O\left( {{X_i},\Theta _c^n} \right) - G_c^n} \right\|}^2}}}{{{C_n}}}} },} 
\end{equation}
where the fix weight $1/C_n$ is used to replace the adaptive weight $W_c^n$ in formulation \ref{e6}. The results are listed in table \ref{table3}. Baseline with channel-wise attention(CA) has achieved 7-point improvement and 7.6-point improvement in MAE and MSE respectively. 
\begin{table}[t]
	\caption{Two sets of ablation experiments of the number of supervisions and the effect of channel-wise attention on ShanghaiTech Part A\cite{MCNN}.}
	\label{table3}
	\begin{center}
		\setlength{\tabcolsep}{4mm}{
			\begin{tabular}{c|cc}
				\toprule[2pt]
				Models & MAE & MSE \\ \midrule[1.5pt]
				Baseline  & 70.2   & 114.6    \\ 
				Baseline-1 & 67.4   & 110.5    \\
				Baseline-2 & 63.9   & 108.2    \\
				Baseline-3 & \text{63.2}   & \textbf{107.0}    \\\midrule[1.5pt]
				Baseline with EW & 68.1 & 111.2 \\ 
				Baseline with CA & \textbf{63.2} &\textbf{107.0}   \\ \bottomrule[2pt]
		\end{tabular}}
	\end{center}
\vspace{-0.5em}
\end{table}
%------------------------------------------------------------------------
\section{Discussion and Analysis}
\subsection{The quality of results}
\label{5.1}

As mentioned above, we regard the decoder of DME as the generation of density maps and conduct supervision to help the generation. It can supplement the feature details lost in the encoder and increase the resolution of result in a more accurate way than interpolation and traditional deconvolution, which improve the quality of generated density maps. To prove the effectiveness in a quantitative way, we employ Peak Signal-to-Noise Ratio (PSNR) and Structural Similarity in Image (SSIM) as the evaluation metrics to evaluate the quality of density maps. Table \ref{table6} shows the results of our MDSNet, baseline and other methods. 

Compared with other methods, the Baseline with MDS achieves a promising performance. Furthermore, the proposed MDS improves the quality of results without increasing the parameters of network.
\begin{table}[t]
	\caption{Evaluation of the quality of results generated by Baseline with/without MDS and other competitive methods on ShanghaiTech Part A\cite{MCNN}.}
	\label{table6}
	\begin{center}  
		\setlength{\tabcolsep}{4mm}{
			\begin{tabular}{c|cc}
				\toprule[2pt]
				Methods       & PSNR  & SSIM \\ \midrule[1.5pt]
				Switching-CNN\cite{Switching-CNN} & 21.91 & 0.67        \\ 
				CP-CNN\cite{CP-CNN}        & 21.72 & 0.72      \\ 
				CSRNet\cite{CSRNet}        & 23.79 & 0.76     \\ 
				Beseline      & 23.53  & 0.76                \\ 
				Baseline with MDS        & \textbf{24.21} &\textbf{0.82}        \\ \bottomrule[2pt]
		\end{tabular}}
	\end{center}
\vspace{-1.5em}
\end{table}

\subsection{Analysis of the Loss Weights}
We analyze the effect of loss weight on the final performance of the model in the Sec.\ref{4.2}. $\alpha$ and ${\beta _n}({\beta_1},{\beta_2},{\beta_3})$ are hyper-parameters in formulation \ref{e6}. In the contrast experiments, we set $alpha$ in \{0.025, 0.05, 0.075, 0.1, 0.125\} to change the balance of loss of groundtruth and loss of features. In addition, we design the other two modes to decide the distribution of ${\beta_n}$: Equal and Decreasing. To be specific, Equal means all features of hidden layers have the same weights and Decreasing means the weight is half of the former layer, which is contrary to Increase. 

\begin{figure}[t]
	\centering
	\includegraphics[scale=0.55]{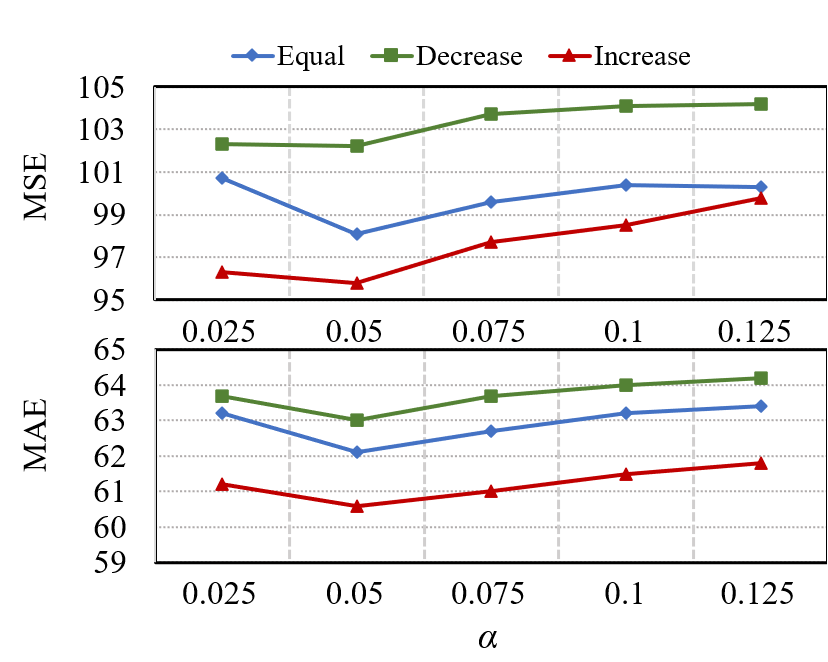}
    \caption{The results of experiments on loss weights.}
	\label{fig5}
\end{figure}

From the Fig.\ref{fig5}, the performance is more sensitive to the distribution of weights. The network achieves lower estimation errors when the distribution of weights follows the Increase mode and the $\alpha$ is set to 0.05.

\subsection{Generalization of MDS}
\label{5.2}
The proposed MDS is a novel framework that is easy to implement on other CNN-based methods. We adapt the MDS to CSRNet and VGGNet by fine-tuning the channels of the generated channels of supervision information to correspond to the decoder. The experimental VGGNet uses the first ten layers of VGG-16 as the encoder and five convolutional layers as the decoder. Table \ref{table4} shows the performance of CSRNet and VGGNet with/without MDS. The performance has improved significantly in terms of estimation errors and the quality of density map with the help of MDS.
\begin{table}[t]
	\caption{Experiments to prove the effectiveness of MDS on ShanghaiTech Part A.}
	\label{table4}
	\begin{center}  
		\setlength{\tabcolsep}{2mm}{
			\begin{tabular}{c|cc|cc}
				\toprule[2pt]
				%	\hline
				\multirow{2}{*}{Methods} & \multicolumn{2}{c|}{VGG-16\cite{VGG}} & \multicolumn{2}{c}{CSRNet\cite{CSRNet}} \\
				& MAE          & MSE          & MAE         & MSE          \\ \midrule[1.5pt]
				Baseline                 & 71.4         & 115.7        & 68.2        & 115.0        \\
				Baseline with MDS        & 65.0         & 113.0        & 63.8        & 109.7        \\ 
				\bottomrule[2pt]
		\end{tabular}}
	\end{center}
\vspace{-1.5em}
\end{table}
%-------------------------------------------------------------------------
\section{Conclusion}
\label{6}
In this paper, we have proposed a novel supervision framework called MDS for crowd counting. It improves the traditional deep supervision methods by conduct supervision on the channels of hidden layers directly. In the proposed MDSNet, we employ an auxiliary network to obtain the information which is conducive to the generation of density maps and the weights of different channels. Then the deep supervision is implemented on several nodes in decoder of DME to supplement the details of feature maps and alleviate the overfitting. Furthermore, the dot annotations are converted to continuous supervision information, which helps us conduct dot supervision directly. Experiments on three datasets prove that the MDS significantly improves the performance of the baseline. Moreover, the results are competitive to state-of-the-art methods in terms of traditional density map supervision and dot supervision.

{\small
	\bibliographystyle{ieee_fullname}
	\bibliography{egbib}

\begin{thebibliography}{10}\itemsep=-1pt

\bibitem{CrowdNet}
L. Boominathan, S. Kruthiventi, and R. Babu.
\newblock Crowdnet: {A} deep convolutional network for dense crowd counting.
\newblock In {\em Proceedings of the ACM International Conference on
  Multimedia}, pages 640--644, 2016.

\bibitem{DSN}
C., S. Xie, P. Gallagher, Z. Zhang, and Z. Tu.
\newblock Deeply-supervised nets.
\newblock In {\em Proceedings of the Society for Artificial Intelligence and
  Statistics}, volume~38, pages 562--570, 2015.

\bibitem{SANet}
X. Cao, Z. Wang, Y. Zhao, and F. Su.
\newblock Scale aggregation network for accurate and efficient crowd counting.
\newblock In {\em Proceedings of the European Conference on Computer Vision},
  volume 11209, pages 757--773, 2018.

\bibitem{ChanLV08}
A. Chan, Z. Liang, and N. Vasconcelos.
\newblock Privacy preserving crowd monitoring: Counting people without people
  models or tracking.
\newblock In {\em Proceedings of the IEEE Conference on Computer Vision and
  Pattern Recognition}, pages 1--7, 2008.

\bibitem{ChenLGX12}
K. Chen, C. Loy, S. Gong, and T. Xiang.
\newblock Feature mining for localised crowd counting.
\newblock In {\em Proceedings of the British Machine Vision Conference}, pages
  1--11, 2012.

\bibitem{2005Histograms}
N. Dalal and B. Triggs.
\newblock Histograms of oriented gradients for human detection.
\newblock In {\em Proceedings of the IEEE Conference on Computer Vision and
  Pattern Recognition}, pages 886--893, 2005.

\bibitem{PCCNet}
J. Gao, Q. Wang, and X. Li.
\newblock {PCC} net: Perspective crowd counting via spatial convolutional
  network.
\newblock {\em IEEE Transactions on Circuits and Systems for Video Technology},
  30(10):3486--3498, 2020.

\bibitem{SEblock}
J. Hu, L. Shen, S. Albanie, G. Sun, and E. Wu.
\newblock Squeeze-and-excitation networks.
\newblock {\em IEEE Transactions on Pattern Analysis and Machine Intelligence},
  42(8):2011--2023, 2020.

\bibitem{Useless2}
S. Huang, X. Li, Z. Zhang, F. Wu, S. Gao, R. Ji, and J. Han.
\newblock Body structure aware deep crowd counting.
\newblock {\em IEEE Transactions on Image Processing}, 27(3):1049--1059, 2018.

\bibitem{UCF_CC_50}
H. Idrees, I. Saleemi, C. Seibert, and M. Shah.
\newblock Multi-source multi-scale counting in extremely dense crowd images.
\newblock In {\em Proceedings of the IEEE Conference on Computer Vision and
  Pattern Recognition}, pages 2547--2554, 2013.

\bibitem{TEDNet}
X. Jiang, Z. Xiao, B. Zhang, X. Zhen, X. Cao, D. Doermann, and L. Shao.
\newblock Crowd counting and density estimation by trellis encoder-decoder
  networks.
\newblock In {\em Proceedings of the IEEE Conference on Computer Vision and
  Pattern Recognition}, pages 6133--6142, 2019.

\bibitem{KangDC16}
D. Kang, D. Dhar, and A. Chan.
\newblock Crowd counting by adapting convolutional neural networks with side
  information.
\newblock {\em arXiv preprint}, arXiv: 1611.06748, 2016.

\bibitem{LempitskyZ10}
V. Lempitsky and A. Zisserman.
\newblock Learning to count objects in images.
\newblock In {\em Proceedings of the Conference and Workshop on Neural
  Information Processing Systems}, pages 1324--1332, 2010.

\bibitem{Related2}
C. Li, M. Zia, Q. Tran, X. Yu, G. Hager, and M. Chandraker.
\newblock Deep supervision with shape concepts for occlusion-aware 3d object
  parsing.
\newblock In {\em Proceedings of the IEEE Conference on Computer Vision and
  Pattern Recognition}, pages 388--397, 2017.

\bibitem{DSPAMI}
C. Li, M. Zia, Q. Tran, X. Yu, G. Hager, and M. Chandraker.
\newblock Deep supervision with intermediate concepts.
\newblock {\em IEEE Transactions on Pattern Analysis and Machine Intelligence},
  41(8):1828--1843, 2019.

\bibitem{extra3}
X. Li, M. Chen, F. Nie, and Q. Wang.
\newblock Locality adaptive discriminant analysis.
\newblock In {\em Proceedings of the International Joint Conference on
  Artificial Intelligence}, pages 2201--2207, 2017.

\bibitem{extra1}
X. Li, M. Chen, F. Nie, and Q. Wang.
\newblock A multiview-based parameter free framework for group detection.
\newblock In {\em Proceedings of the Thirty-First AAAI Conference on Artificial
  Intelligence}, pages 4147--4153, 2017.

\bibitem{CSRNet}
Y. Li, X. Zhang, and D. Chen.
\newblock Csrnet: Dilated convolutional neural networks for understanding the
  highly congested scenes.
\newblock In {\em Proceedings of the IEEE Conference on Computer Vision and
  Pattern Recognition}, pages 1091--1100, 2018.

\bibitem{coushu7}
Z. Li, Z. Chen, Q. Wu, and C. Liu.
\newblock Real-time pedestrian detection with deep supervision in the wild.
\newblock {\em Signal Image and Video Process}, 13(4):761--769, 2019.

\bibitem{DecideNet}
J. Liu, C. Gao, D. Meng, and A. Hauptmann.
\newblock Decidenet: Counting varying density crowds through attention guided
  detection and density estimation.
\newblock In {\em Proceedings of the IEEE Conference on Computer Vision and
  Pattern Recognition}, pages 5197--5206, 2018.

\bibitem{coushu2}
L. Liu, H. Wang, G. Li, W. Ouyang, and L. Lin.
\newblock Crowd counting using deep recurrent spatial-aware network.
\newblock In {\em Proceedings of the International Joint Conference on
  Artificial Intelligence}, pages 849--855, 2018.

\bibitem{CAN}
W. Liu, M. Salzmann, and P. Fua.
\newblock Context-aware crowd counting.
\newblock In {\em Proceedings of the IEEE Conference on Computer Vision and
  Pattern Recognition}, pages 5099--5108, 2019.

\bibitem{2006Detecting}
X. Liu, P. Tu, J. Rittscher, A. Perera, and N. Krahnstoever.
\newblock Detecting and counting people in surveillance applications.
\newblock In {\em Proceedings of the IEEE International Conference on Advanced
  Video and Signal-based Surveillance}, pages 306--311, 2005.

\bibitem{Self-supervision}
X. Liu, J. Weijer, and A. Bagdanov.
\newblock Leveraging unlabeled data for crowd counting by learning to rank.
\newblock In {\em Proceedings of the IEEE Conference on Computer Vision and
  Pattern Recognition}, pages 7661--7669, 2018.

\bibitem{coushu8}
Y. Liu and M. Lew.
\newblock Learning relaxed deep supervision for better edge detection.
\newblock In {\em Proceedings of the IEEE Conference on Computer Vision and
  Pattern Recognition}, pages 231--240, 2016.

\bibitem{coushu6}
A. Luo, F. Yang, X. Li, D. Nie, Z. Jiao, S. Zhou, and H. Cheng.
\newblock Hybrid graph neural networks for crowd counting.
\newblock In {\em Proceedings of the Thirty-Fourth AAAI Conference on
  Artificial Intelligence}, pages 11693--11700, 2020.

\bibitem{Bayesian}
Z. Ma, X. Wei, X. Hong, and Y. Gong.
\newblock Bayesian loss for crowd count estimation with point supervision.
\newblock In {\em Proceedings of the International Conference on Computer
  Vision}, pages 6141--6150, 2019.

\bibitem{DUBNet}
M. Oh, P. Olsen, and K. Ramamurthy.
\newblock Crowd counting with decomposed uncertainty.
\newblock In {\em Proceedings of the Thirty-Fourth AAAI Conference on
  Artificial Intelligence}, pages 11799--11806, 2020.

\bibitem{Regression1}
T. Ojala and M. inen.
\newblock Multiresolution gray-scale and rotation invariant texture
  classification with local binary patterns.
\newblock {\em IEEE Transactions on Pattern Analysis and Machine Intelligence},
  24(7):971--987, 2002.

\bibitem{D4}
J. Rittscher, P. Tu, and N. Krahnstoever.
\newblock Simultaneous estimation of segmentation and shape.
\newblock In {\em Proceedings of the IEEE Conference on Computer Vision and
  Pattern Recognition}, pages 486--493, 2005.

\bibitem{Useless1}
D. Rubio and R. Sastre.
\newblock Towards perspective-free object counting with deep learning.
\newblock In {\em Proceedings of the European Conference on Computer Vision},
  volume 9911, pages 615--629, 2016.

\bibitem{coushu1}
D. Sam and R. Babu.
\newblock Top-down feedback for crowd counting convolutional neural network.
\newblock In {\em Proceedings of the Thirty-Second AAAI Conference on
  Artificial Intelligence}, pages 7323--7330, 2018.

\bibitem{coushu3}
D. Sam, N. Sajjan, R. Babu, and M. Srinivasan.
\newblock Divide and grow: Capturing huge diversity in crowd images with
  incrementally growing {CNN}.
\newblock In {\em Proceedings of the IEEE Conference on Computer Vision and
  Pattern Recognition}, pages 3618--3626, 2018.

\bibitem{coushu4}
D. Sam, N. Sajjan, H. Maurya, and R. Babu.
\newblock Almost unsupervised learning for dense crowd counting.
\newblock In {\em {AAAI}}, pages 8868--8875, 2019.

\bibitem{Switching-CNN}
D. Sam, S. Surya, and R. Babu.
\newblock Switching convolutional neural network for crowd counting.
\newblock In {\em Proceedings of the IEEE Conference on Computer Vision and
  Pattern Recognition}, pages 4031--4039, 2017.

\bibitem{True3}
Z. Shen, Z. Liu, J. Li, Y. Jiang, Y. Chen, and X. Xue.
\newblock Object detection from scratch with deep supervision.
\newblock {\em IEEE Transactions on Pattern Analysis and Machine Intelligence},
  42(2):398--412, 2020.

\bibitem{ACSCP}
Z. Shen, Y. Xu, B. Ni, M. Wang, J. Hu, and X. Yang.
\newblock Crowd counting via adversarial cross-scale consistency pursuit.
\newblock In {\em Proceedings of the IEEE Conference on Computer Vision and
  Pattern Recognition}, pages 5245--5254, 2018.

\bibitem{VGG}
K. Simonyan and A. Zisserman.
\newblock Very deep convolutional networks for large-scale image recognition.
\newblock volume arXiv: 1409.1556, 2014.

\bibitem{2017A}
V. Sindagi and M. Patel.
\newblock A survey of recent advances in cnn-based single image crowd counting
  and density estimation.
\newblock {\em Pattern Recognition Letters}, 107:3--16, 2018.

\bibitem{CMTL}
V. Sindagi and V. Patel.
\newblock Cnn-based cascaded multi-task learning of high-level prior and
  density estimation for crowd counting.
\newblock In {\em Proceedings of the IEEE International Conference on Advanced
  Video and Signal-based Surveillance}, pages 1--6, 2017.

\bibitem{CP-CNN}
V. Sindagi and V. Patel.
\newblock Generating high-quality crowd density maps using contextual pyramid
  cnns.
\newblock In {\em Proceedings of the International Conference on Computer
  Vision}, pages 1879--1888, 2017.

\bibitem{GoogleNet}
C. Szegedy, W. Liu, Y. Jia, P. Sermanet, S. Reed, D. Anguelov, D. Erhan, V.
  Vanhoucke, and A. Rabinovich.
\newblock Going deeper with convolutions.
\newblock In {\em Proceedings of the IEEE Conference on Computer Vision and
  Pattern Recognition}, pages 1--9, 2015.

\bibitem{Wan}
J. Wan and A. Chan.
\newblock Adaptive density map generation for crowd counting.
\newblock In {\em Proceedings of the International Conference on Computer
  Vision}, pages 1130--1139, 2019.

\bibitem{WangZYLC15}
C. Wang, H. Zhang, L. Yang, S Liu, and X. Cao.
\newblock Deep people counting in extremely dense crowds.
\newblock In {\em Proceedings of the ACM International Conference on
  Multimedia}, pages 1299--1302, 2015.

\bibitem{NWPUdataset}
Q. Wang, J. Gao, W. Lin, and X. Li.
\newblock Nwpu-crowd: A large-scale benchmark for crowd counting and
  localization.
\newblock {\em IEEE Transactions on Pattern Analysis and Machine Intelligence},
  PP(99):1--1, 2020.

\bibitem{GTA}
Q. Wang, J. Gao, W. Lin, and Y. Yuan.
\newblock Learning from synthetic data for crowd counting in the wild.
\newblock In {\em Proceedings of the IEEE Conference on Computer Vision and
  Pattern Recognition}, pages 8198--8207, 2019.

\bibitem{IJCV}
Q. Wang, J. Gao, W. Lin, and Y. Yuan.
\newblock Pixel-wise crowd understanding via synthetic data.
\newblock {\em International Journal of Computer Vision}, 2020.

\bibitem{UCF-QNRF}
X. Wang, T. Xiao, Y. Jiang, S. Shao, J. Sun, and C. Shen.
\newblock Repulsion loss: Detecting pedestrians in a crowd.
\newblock In {\em Proceedings of the European Conference on Computer Vision},
  pages 7774--7783, 2018.

\bibitem{2005Detection}
B. Wu and R. Nevatia.
\newblock Detection of multiple, partially occluded humans in a single image by
  bayesian combination of edgelet part detectors.
\newblock In {\em Proceedings of the International Conference on Computer
  Vision}, pages 90--97, 2005.

\bibitem{DSN2}
S. Xie and Z. Tu.
\newblock Holistically-nested edge detection.
\newblock In {\em Proceedings of the International Conference on Computer
  Vision}, pages 1395--1403, 2015.

\bibitem{Cross-scene}
C. Zhang, H. Li, X. Wang, and X. Yang.
\newblock Cross-scene crowd counting via deep convolutional neural networks.
\newblock In {\em Proceedings of the IEEE Conference on Computer Vision and
  Pattern Recognition}, pages 833--841, 2015.

\bibitem{MCNN}
Y. Zhang, D. Zhou, S. Chen, S. Gao, and Y. Ma.
\newblock Single-image crowd counting via multi-column convolutional neural
  network.
\newblock In {\em Proceedings of the IEEE Conference on Computer Vision and
  Pattern Recognition}, pages 589--597, 2016.

\bibitem{True4}
H. Zhao, J. Shi, X. Qi, X. Wang, and J. Jia.
\newblock Pyramid scene parsing network.
\newblock In {\em Proceedings of the IEEE Conference on Computer Vision and
  Pattern Recognition}, pages 6230--6239, 2017.

\bibitem{Dectect3}
T. Zhao and R. Nevatia.
\newblock Tracking multiple humans in complex situations.
\newblock {\em IEEE Transactions on Pattern Analysis and Machine Intelligence},
  26(9):1208--1221, 2004.

\bibitem{coushu5}
Z. Zhao, M. Shi, X. Zhao, and L. Li.
\newblock Active crowd counting with limited supervision.
\newblock In {\em Proceedings of the European Conference on Computer Vision},
  volume 12365, pages 565--581, 2020.

\end{thebibliography}
}
\end{document}